\documentclass[a4paper]{article}

\usepackage{INTERSPEECH2019}
\usepackage{cite}
\usepackage{amsmath,amssymb}
\DeclareMathOperator{\E}{\mathbb{E}}
\begin{document}
\title{SPEAK YOUR MIND! \\Towards Imagined Speech Recognition With Hierarchical Deep Learning}
\name{Pramit Saha$^1$, Muhammad Abdul-Mageed$^2$, Sidney Fels$^1$}
\address{
  $^1$Human Communication Technologies Lab, University of British Columbia
  $^2$Natural Language Processing Lab, University of British Columbia
  }

\email{pramit@ece.ubc.ca, muhammad.mageed@ubc.ca, ssfels@ece.ubc.ca}
\maketitle
\begin{abstract}
Speech-related Brain Computer Interface (BCI) technologies provide effective vocal communication strategies for controlling devices through speech commands interpreted from brain signals.  In order to infer imagined speech from active thoughts, we propose a novel hierarchical deep learning BCI system for subject-independent classification of 11 speech tokens including phonemes and words. Our novel approach exploits predicted articulatory information of six phonological categories (e.g., nasal, bilabial) as an intermediate step for classifying the phonemes and words, thereby finding discriminative signal responsible for natural speech synthesis.  The proposed network is composed of hierarchical combination of spatial and temporal CNN  cascaded with a deep autoencoder. Our best models on the KARA database achieve an average accuracy of 83.42\% across the six different binary phonological classification tasks, and 53.36\% for the individual token identification task, significantly outperforming our baselines. Ultimately, our work suggests the possible existence of a brain imagery footprint for the underlying articulatory movement related to different sounds that can be used to aid imagined speech decoding. 
 
\end{abstract}


\section{Introduction}
Speech-related Brain Computer Interface (BCI) technologies provide neuro-prosthetic help for people with speaking disabilities, neuro-muscular disorders and diseases. It can equip these users with a medium to communicate and express their thoughts, thereby improving the quality of rehabilitation and clinical neurology. Such devices also have applications for healthy individuals--in entertainment, preventive treatments, personal communication, games, etc. 

Typical forms of daily human interaction involve verbal and non-verbal communication in the form of vocal speech (or sounds) and physical gestures. However, the majority of existing research focuses on motor imagery-based control of external devices \cite{pfurtscheller2001motor,herman2008comparative} (e.g., wheelchair). For communicating expressions and thoughts, we need a control space equipped with more functionalities and higher degrees of freedom. For these reasons, the vocal space involving labial, lingual, naso-pharyngeal and jaw motion is arguably an alternative, multi-dimensional controlling paradigm.

The challenge is that speech production is a complex process, involving intricate muscular hydrostat structure movement (e.g., the tongue). Recently, deep neural networks have emerged as efficient tools for handling complex tasks. Yet, there is hardly any work investigating the applicability and performance of such deep learning techniques for speech imagery-based BCI. 

Among the various brain activity-monitoring modalities in BCI, Electroencephalography (EEG) \cite{machado2010eeg,lotte2007review} has been demonstrated as carrying promising signal for differentiating different brain activities (through measurement of related electric fields). However, these are high dimensional, and have poor Signal-to-Noise ratio, low spatial resolution, and plenty of artifacts. Besides, it is not entirely clear how to decode the desired information from the high-dimensional raw EEG signals. 
\begin{figure}[t]
\centering
\includegraphics[width=8.5cm,height=9cm,keepaspectratio]{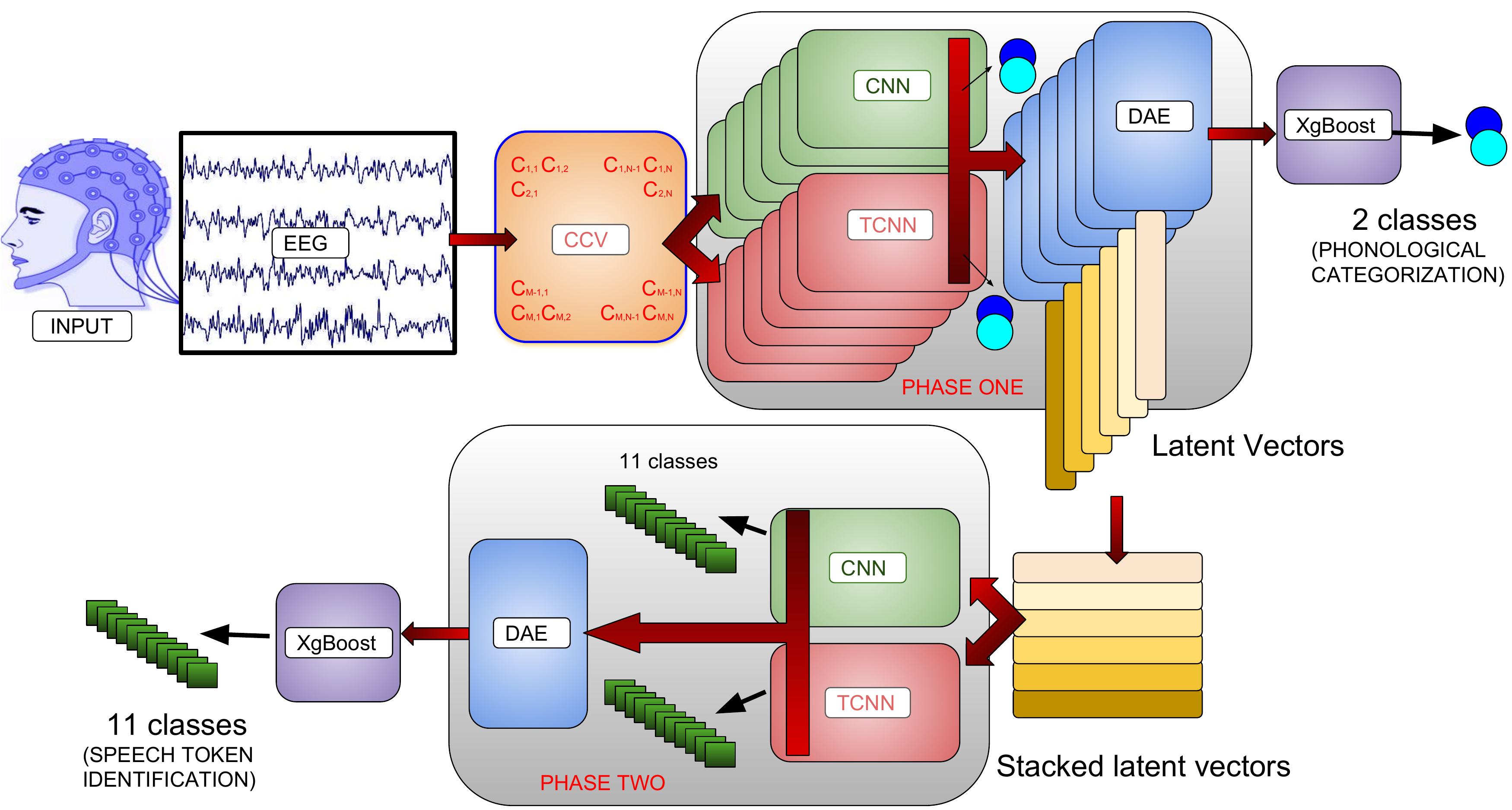}
\caption{Overall framework of the proposed approach}
\label{fig:frameworkph2}
\end{figure}

Although the area of BCI based speech intent recognition has received increasing attention within the research community in the past few years, most research has focused on classification of individual speech categories in terms of discrete vowels, phonemes and words  \cite{dasalla2009single,dasalla2009spatial,idrees2016vowel,deng2010eeg, kim2014eeg,brigham2010imagined,mohanchandra2016communication,wang2013analysis,gonzalez2017sonification}. This includes categorization of imagined EEG signal into binary vowel categories like \textit{/a/}, \textit{/u/} and rest \cite{dasalla2009single,dasalla2009spatial,idrees2016vowel}; binary syllable classes like \textit{/ba/} and \textit{/ku/}~\cite{d2009toward,deng2010eeg,kim2014eeg,brigham2010imagined}; a handful of control words like \textit{'up'}, \textit{'down'}, \textit{'left'}, \textit{'right'} and \textit{'select'} \cite{gonzalez2017sonification} or others like \textit{'water'}, \textit{'help'}, \textit{'thanks'}, \textit{'food'}, \textit{'stop'} ~\cite{mohanchandra2016communication}, Chinese characters~\cite{wang2013analysis}, etc. Such works mostly involve traditional signal processing or manual feature handcrafting along with linear classifiers (e.g., SVMs). In our recent work\cite{saha2018hierarchical}, we introduced deep learning models for classification of vowels and words that achieve 23.45\% improvement of accuracy over the baseline.

In this work, our goal is to detect speech tokens from speech imagery (\textit{active thoughts} or \textit{imagined speech} \cite{wang2013analysis}). Speech imagery is about representing speech in terms of sounds inside the human brain without overt vocalization nor articulatory movements. We hypothesize the existence of some sort of brain footprint for articulatory movements underlying related speech token imagery. Hence, we attempt to first predict phonological categories and then use these predictions to aid recognition of imagined speech at the token level (phonemes and words). We introduce our framework for solving this problem next.


\section{Proposed Deep Learning Framework}
\subsection{Mathematical Formulation}
We denote the multivariate time-series data as $ X \in ~R^C*T $ , with sets of labels $Y \in {y_{1},y_{2}, ..., y_{11}}$ where $X$ corresponds to the single trial EEG data, having a number of channels $C$, and for a number of time steps $T$. $Y$ is a one-hot encoded vector of 11 labels corresponding to individual words and labels. In our case, $C$ is 64 and time interval is represented in terms of 5,000 time steps. As discussed earlier, we essentially build our system in two consequent steps: 
The first step is binary classification of $ X \in ~R^C*T $ into presence or absence of 6 phonological categories:   \{$ z_{1}, \Bar{z_{1}} \}  $, \{$ z_{2}, \Bar{z_{2}}$\}, \{$ z_{3}, \Bar{z_{3}}$\}, \{$ z_{4}, \Bar{z_{4}}$\}, \{$ z_{5}, \Bar{z_{5}}$\}, \{$ z_{6}, \Bar{z_{6}}$\}. The second step is tease apart the concatenated autoencoder latent vectors from these 6 classification models \textit{viz.}, $W=
\bigcup_{i=1}^{6} w_{i}$ into 11 classes: \{${y_{1},y_{2}, ..., y_{11}}$\} where $w_{i}$ corresponds to latent vector space corresponding to $i^{th}$ phonological classification into \{$ z_{i}, \Bar{z}_{i} $\}.
\subsection{Predicting Phonological Categories}

We build on our hypothesis that the active thought process underlying covert speech does have some relevant features corresponding to the intended activity of nasopharynx, lips, tongue movements and positions etc. Hence, in the first phase, we target five binary classification tasks addressed in ~\cite{zhao2015classifying,sun2016neural}, \textit{i.e.} presence/absence of consonants, phonemic nasal, bilabial, high-front vowels and high-back vowels. Additionally, we add a voiced vs. voiceless classification task whose goal is to provide information about the intended involvement of vocal folds. In this way, rather than directly discriminating the individual phonemes and words, we first attempt to accurately classify imagined phonological categories on the basis of underlying intended articulatory movements. 

Rather than using the raw multi-channel high-dimensional EEG data (which requires long training times and resources), we experimentally\footnote{We do not report these experiments here, for space limitation.} found that it is a better strategy to first reduce the dimensionality of the EEG by capturing the joint variability of the electrodes. Crucially, our target was to model the directional relationship and dependency among the electrodes over the entire time interval. Hence, instead of the conventional approach of selecting a handful of channels as in ~\cite{zhao2015classifying,sun2016neural}, we address this issue by computing the channel cross-covariance (CCV), resulting in positive, semi-definite matrices encoding the connectivity of the electrodes. We define CCV between any two electrodes $c_1$ and $c_2$ as:
$Cov({X}^{c_{1}}_t,{X}^{c_{2}}_{t+\tau})=\displaystyle \E[X^{c_{1}}(t)-\mu_{X^{c_{1}}}(t)][X^{c_{2}}(t+\tau)-\mu_{X^{c_{2}}}(t+\tau)]$.

We use convolutional neural networks (CNNs) \cite{krizhevsky2012imagenet} to extract the spatial features from the covariance matrix. Each layer decodes non-linear spatial feature representations from the previous layer using convolutional filters and non-linear ReLU \cite{nair2010rectified} activation functions applied to the resulting feature maps. We employ a four-layered 2D CNN stacking two convolutional and two fully connected hidden layers. This is the first level of hierarchy where the network is trained with the corresponding labels as target outputs, optimizing cross-entropy cost function. We describe architectural and hyper-parameter choices for our networks in Table ~\ref{hyp}.

In parallel with CNN, we apply a temporal CNN (TCNN) \cite{van2016wavenet, bai2018empirical} on the channel covariance matrices to explore the hidden temporal features of the electrodes. Namely, we flatten the lower triangular matrix of the CCV and feed the data of length 1,891 to the TCNN. In order to capture the long term dependencies and temporal correlations of the signal, we exploit a 6 layer stacked TCNN and train in a similar manner as CNN, using Adam \cite{kingma2014adam} to optimize cross-entropy function. We use stacked dilation filters with a dilation factor of 2, resulting in exponential growth of receptive field with depth and increase in model capacity. This essentially enhances the non-linear discriminative power of the network, which is vital for our problem space. 
\begin{figure}
\centering
\includegraphics[width=15cm,height=6cm,keepaspectratio]{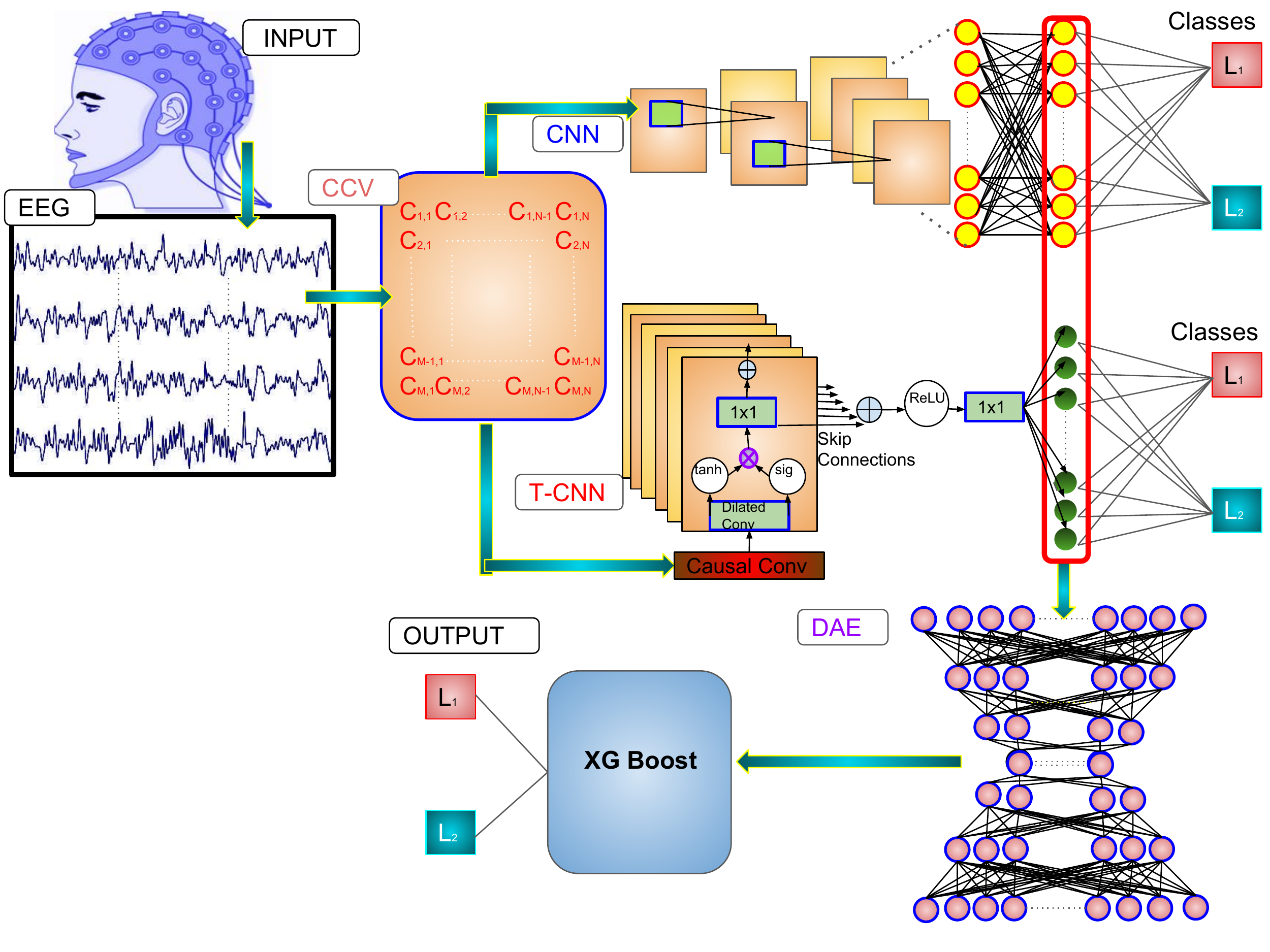}
\caption{Overview of phonological prediction of our novel architecture}
\label{fig:frameworkph1}
\end{figure}
We concatenate the last fully-connected layer from the CNN with its counterpart in the TCNN to compose a single feature vector based on these two penultimate layers thereby forming a joint spatio-temporal encoding of the cross-covariance matrix. In order to further reduce the dimensionality of the spatio-temporal encodings and cancel background noise effects ~\cite{zhang2018converting}, we train an unsupervised deep autoenoder (DAE) \cite{goodfellow2016deep} on the fused heterogeneous features produced by the combined CNN and TCNN information. The DAE forms our second level of hierarchy, with 3 encoding and 3 decoding layers, and mean squared error (MSE) as the cost function. 

At the third level of hierarchy, the discrete latent vector representation of the deep autoencoder is fed into an Extreme Gradient Boost based classification layer ~\cite{chen2015xgboost,chen2016xgboost} motivated by ~\cite{zhang2018converting}. The classifier receives its input from the latent vectors of the deep autoencoder and is trained in a supervised manner to output the final predicted phonological classes corresponding to speech imagery. 

\subsection{Predicting Speech Tokens}
Next, our goal is to use the combined information available from all the six phonological categories to predict the 11 individual speech tokens present in our EEG dataset (introduced in Section \ref{subsec:dataset}). Such a hierarchical approach essentially differs from the direct speech classification approach as it imposes richer constraints on the information space by involving features from all the phonological categorization tasks. Our results show the utility of this approach as we report in Section \ref{subsec:phological_res}. To this end, we first stack the bottleneck features of the autoencoders corresponding to the aforementioned six classification tasks, into a matrix of dimensions $6\times 256$. In order to explicitly exploit phonological information in the imagined speech recognition task, we feed this stacked latent matrix as the input to our classification model similar to the first phase. 
\section{Experiments}
\subsection{Dataset}\label{subsec:dataset}
We evaluate our models on a publicly available dataset, KARA ONE ~\cite{zhao2015classifying}. It is composed of multimodal data for stimulus-based, imagined and articulated speech state corresponding to 7 phonemic/syllabic ( \texttt{/iy/}, \texttt{/piy/}, \texttt{/tiy/}, \texttt{/diy/},  \texttt{/uw/}, \texttt{/m/}, \texttt{/n/}) as well as 4 words (\texttt{pat}, \texttt{pot}, \texttt{knew} and \texttt{gnaw}). The study comprising the dataset consists of 14 participants, with each prompt presented 11 times to each individual. Since our intention is to classify the phonological categories from human thoughts, we discard the facial and audio information and only consider the EEG data corresponding to imagined speech. 
More details regarding the database can be found in ~\cite{zhao2015classifying}.
\subsection{Procedure and Model Training}

We randomly shuffle and divide the data (1913 signals from 14 individuals) into train (80\%), development (10\%) and test sets (10\%). The architectural parameters and hyperparameters listed in Table \ref{hyp} were selected through an exhaustive grid-search based on the development set. We conduct a series of empirical studies starting from single hidden-layered networks for each of the blocks and, based on the validation accuracy, we increase the depth of each given network and select the optimal parametric set from all possible combinations of parameters. For the gradient boosting classification, we fix the maximum depth at 10, number of estimators at 5,000, learning rate at 0.1,  regularization coefficient at 0.3, subsample ratio at 0.8, and column-sample/iteration at 0.4. We did not find any notable change of accuracy while varying other hyperparameters while training gradient boost classifier. For the phonological categorization task, input data for CNN and TCNN (covariance matrix) is of length $61 \times 61$ and 1,891 respectively, while for the speech recognition task, the input data (phonological features) is of length $6 \times 256$ and 1,536 respectively. The input data for deep autoencoders pertaining the two tasks is of length 2,915 (1,891 TCNN + 1,024 CNN features). 
\begin{table}[]
\scriptsize 
 \caption{Selected parameter sets}
\label{hyp}
\begin{tabular}{*{5}{p{1.55cm}}}
\hline\hline
\textbf{Parameters}&\textbf{CNN}  & \textbf{TCNN}& \textbf{DAE}\\ \hline\hline
Epochs    &50     & 50  & 200   \\ \hline
Total layers & 6  & 6 & 7 \\ \hline 
Hidden layers' details    & Conv:32,64 masks:3x3 Dense: 64,128  &  mask: 5, Dilation : 2 &  E:1024,512,128 
D:128,512,1024 \\ \hline
Activations &  ReLU, last-layer : softmax & sigm, tanh, ReLU, last-layer : softmax & ReLU, ReLU, sigm, sigm, ReLU, tanh \\ \hline
Dropout & .25,~~ .50 & .25,~~ .50 & .25,~.25,~ .25
\\ \hline
Optimizer & Adam & Adam & Adam
\\ \hline
Loss & Categorical cross entropy & Categorical cross entropy & Mean Sq Error
\\ \hline
l-rate & .001 & .002 & .001
\\ \hline\hline
\end{tabular}
\end{table}
\subsection{Baselines}
We use two baselines, one based on an individual LSTM and another based on an individual CNN. In each case, we pass the data from the cross-variance matrix and classify directly based on output from each of these networks. In addition, we compare to previous works on the same dataset \cite{zhao2015classifying,sun2016neural}. For meaningful comparisons, since these previous works follow a cross-validation set up (14-fold where the model is trained on 13 subjects' data and tested on the $14^{th}$), we mimic the same data splits and report accuracy. To establish a benchmark for computationally costly deep learning work, we choose our 80\%, 10\%, 10\% data splits after shuffling the data. 
\subsection{Results of phonological category prediction}\label{subsec:phological_res}
\begin{table}
\centering
\scriptsize 
 \caption{Results in accuracy on 10\% test data for phonological prediction. \textbf{C-L-D:} CNN+LSTM+DAE}
\label{test_result1}
\begin{tabular}{lllllll}   
\hline
\textbf{Method}  &$\pm$ \textbf{Bilab}& $\pm$ \textbf{Nasal}& \textbf{C/V }&$\pm$ \textbf{/uw/}&$\pm$ \textbf{/iy/ } & \textbf{Avg }\\ \hline
LSTM   & 46.07  & 45.31   & 45.83   & 48.44  & 46.88& 46.51     \\ 
CNN    &59.16     & 57.20   & 67.88    & 69.56    & 68.60 &   64.48  \\ 
CNN+LSTM      & 62.03  & 60.89 & 70.04 & 72.76 &  63.75 & 65.89\\ 
C-L-D        & 78.65  & 74.57  & 87.96 & 83.25 & 77.30  & 80.35 \\
Our model   &   \textbf{81.67} & \textbf{78.33} & \textbf{89.16} & \textbf{85.00} & \textbf{87.20} &\textbf{ 84.27} \\  \hline

\end{tabular}
\end{table}

\begin{table}
\centering
\scriptsize 
 \caption{Classification Performance metrics on 10\% test data in phonological prediction task}
\label{test_result1_1}
\begin{tabular}{llllll}   
\hline
\textbf{Metrics}  &\textbf{Precision}& \textbf{Recall}& \textbf{Specificity}& \textbf{f1 score} & \textbf{Kappa} \\ \hline
  $\pm$ \textbf{Bilab} & 72.09&	75.61&	84.81&	73.81&	63.34   \\ 
   $\pm$ \textbf{Nasal} &67.44&	70.73&	82.28&	69.05&	56.66    \\ 
   ~~~~\textbf{C/V }   & 86.36&	65.52&	96.7&	74.51&	78.32 \\ 
  $\pm$ \textbf{/uw/}  & 77.27&	56.67&	94.44&	65.39&	70.00  \\
 $\pm$ \textbf{/iy/ }  &  86.04	& 78.72	& 91.78	& 82.22	& 74.40 \\                              
$\pm$ \textbf{Voiced} & 78.95 &	86.96 &	68.63 &	82.76 &	58.32\\ \hline

\end{tabular}
\end{table}
\begin{figure}[]
\centering
\includegraphics[width=8cm,height=4cm,keepaspectratio]{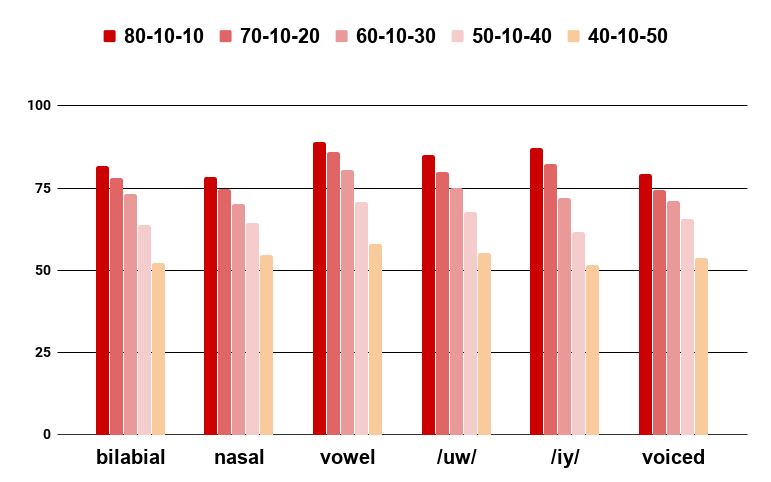}
\caption{Variation of performance accuracy of phonological prediction with varying training-validation-test data ratio}
\label{fig:datasplitph1}
\end{figure}

To demonstrate the significance of the hierarchical CNN-TCNN-DAE method, we also conduct separate experiments with the individual networks and summarize the results in Table \ref{test_result1}. From the average accuracy scores, we observe that our proposed network performs much better than individual networks. A detailed analysis on repeated runs further shows that in most of the cases, LSTM alone does not perform better than chance. CNN, on the other hand, is heavily biased towards the class label from which it sees more training data. Although the situation improves with combined CNN-LSTM, our analysis clearly shows the necessity of a better encoding scheme to utilize the combined features rather than mere concatenation of the penultimate features of both networks. CNN-LSTM-DAE improves classification accuracy by a significant margin, thus demonstrating the utility of the autoencoder contribution towards filtering out the unrelated and noisy features from the concatenated penultimate feature set. 
Replacing the LSTM block with TCNN block endows the network with more temporal discriminative power, resulting in an increase of 3.93\% mean accuracy as shown in Table \ref{test_result1}.
In addition to accuracy, we provide the precision, recall, specificity, f1 score and Kappa coefficients of our method for all the six classification tasks in Table \ref{test_result1_1}. Kappa coefficients offer a metric for evaluating the utility of classifier decisions beyond mere chance ~\cite{tabar2016novel}. Here, a higher mean kappa value corresponding to a task implies that the network is able to find better discriminative information from the EEG data beyond random decisions. The maximum above-chance accuracy (78.32\%) is recorded for presence/absence of the vowel task and the minimum (56.66\%) is recorded for the $\pm{nasal}$. 

Further, to evaluate the robustness of our model against availability of data, we run a set of experiments varying the train-test ratio of the data (results shown in Figure \ref{fig:datasplitph1}). As Figure \ref{fig:datasplitph1} shows, even with less training data (40\% ) and more, and potentially more diverse test data (50\%), our model performs above chance, which indicates its reliability even under these extreme data distribution condition.

\begin{table}[b]
\centering
\scriptsize
  \caption{Comparison in accuracy with Z\_R: ~\cite{zhao2015classifying} and S\_Q: ~\cite{sun2016neural}}
  \label{tab:accuracy}
  \centering
  \begin{tabular}{lllllll}
    \hline
    &$\pm$ \textbf{Bilabial}& $\pm$\textbf{ Nasal}& \textbf{C/V} &$\pm$ \textbf{/uw/}&$\pm$ \textbf{/iy/}\\\hline
    Z\_R &56.64&63.5&18.08&79.16&59.6\\ 
    S\_Q &53&47&25&74&53\\
    Ours &\textbf{75.55}& \textbf{73.45}&\textbf{	85.23}&\textbf{	81.99}&	\textbf{73.30}\\\hline
\end{tabular}
\end{table}

We next compare our phonological prediction to ~\cite{zhao2015classifying} and ~\cite{sun2016neural}. As shown in Table \ref{tab:accuracy}, since the model encounters the unseen data of a new subject for testing, and given the high inter-subject variability of the EEG data, a reduction in the accuracy is expected. However, our network still manages to achieve an improvement of \textbf{18.91},\textbf{ 9.95}, \textbf{67.15}, \textbf{2.83} and \textbf{13.70 \%} over \cite{zhao2015classifying}. Besides, our best model shows more reliability compared to previous works: The standard deviation of our model's classification accuracy across all the tasks is reduced from 22.59\% \cite{zhao2015classifying} and 17.52\%\cite{sun2016neural} to a mere 5.41\%.

\subsection{Results of speech token prediction}

\begin{table}
\centering
\scriptsize 
 \caption{Comparison of accuracy on 10\% test data for speech token prediction task}
\label{test_result2}
\begin{tabular}{lcc}   
\hline
\textbf{Method}  &\textbf{EEG data}&\textbf{Phonological features}\\ \hline
LSTM   & 8.45 &	15.83    \\ 
CNN    & 8.88 &	16.02  \\ 
CNN+LSTM      & 12.44 &	22.10 \\ 
CNN+LSTM+DAE   &  23.45 &	49.19 \\
Our model   &   \textbf{28.08} &	\textbf{53.36}  \\                              \hline

\end{tabular}
\end{table}
\begin{figure}
\centering
\includegraphics[width=8cm,height=10cm,keepaspectratio]{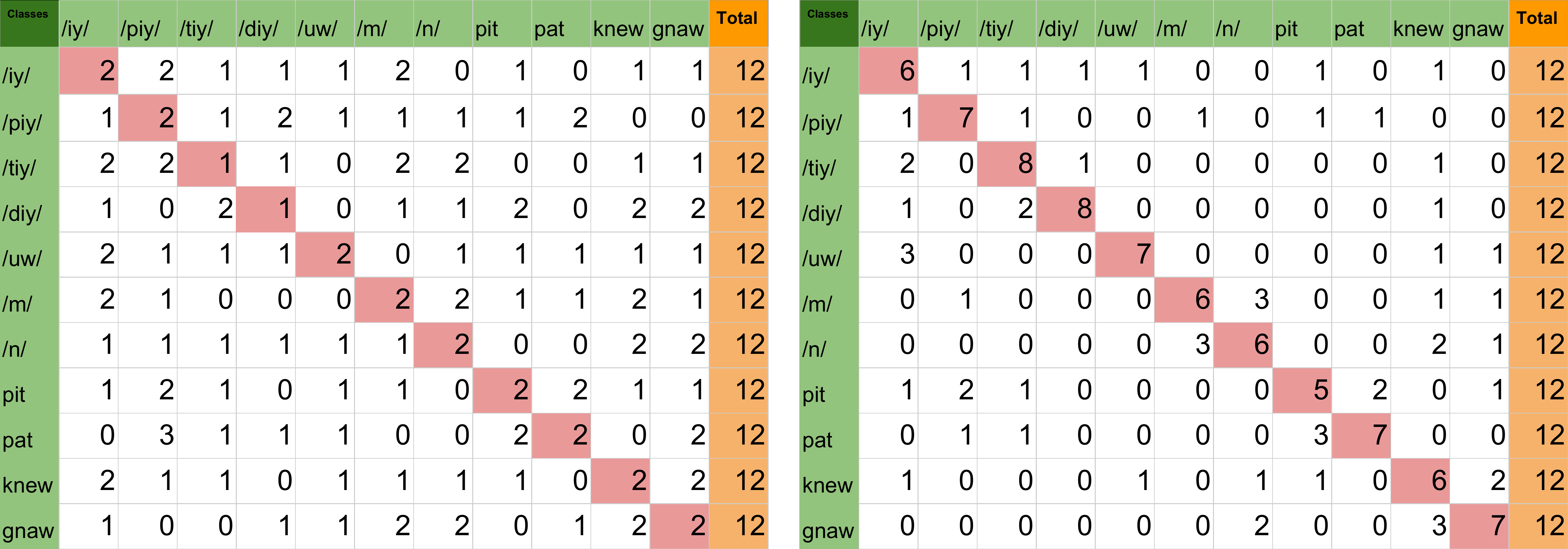}
\caption{Inter-subject confusion matrix for speech token prediction with covariance data (left) and with  phonological feature data (right)}
\label{fig:confmatrix}
\end{figure}
\begin{figure}
\centering
\includegraphics[width=6cm,height=8cm,keepaspectratio]{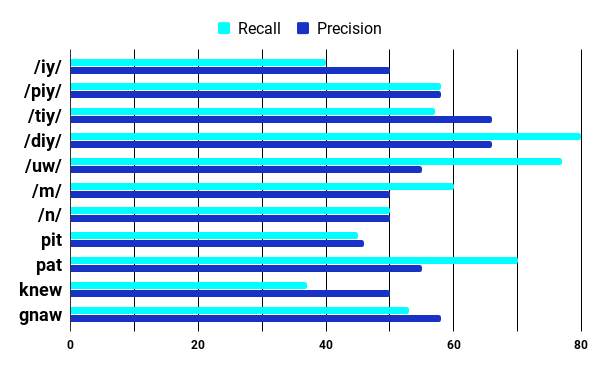}
\caption{Precision and recall metrics corresponding to each speech token on 10\% train data}
\label{fig:recall}
\end{figure}

\begin{figure}
\centering
\includegraphics[width=6cm,height=8cm,keepaspectratio]{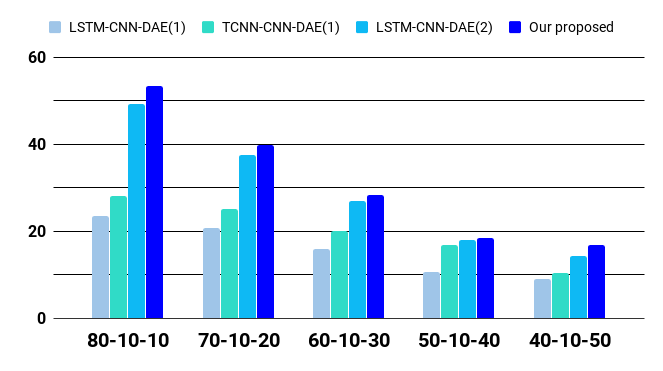}
\caption{Variation of performance accuracy of speech token prediction for top 4 algorithms with varying training-validation-test data ratio}
\label{fig:datasplitph2}
\end{figure}

We provide performance of the baseline methods on direct covariance data and phonological feature data in Table \ref{test_result2}. For a closer look at the results, we report sample confusion matrix of our model on a leave-one-subject-out classification strategy in Figure \ref{fig:confmatrix}. In this step, we essentially train the network on the data of 13 subjects and test on the $14^{th}$ subject, to check the inter-subject variability of our model. As it is evident from the figure, with direct covariance data, the predicted classes corresponding to each true label are widely distributed throughout the matrix and hardly gives any significant information about the actual speech token. However, involvement of the phonological categorization as an intermediate step increases the prediction accuracy. Interestingly, the false negatives corresponding to each of the tokens also inform us about the respective structure of the word or phoneme. For example, the misclassification of \texttt{/n/} as \texttt{/m/}, \texttt{`knew'} and \texttt{`gnaw'} in a few cases, show that while the network gets strong discriminative features from the other five networks, features pertaining to the nasal category require more discriminative ability to more accurately categorize the phoneme \texttt{/n/}. Such an observation indeed proves that the phonological features play a significant role for achieving an accurate classification of the speech tokens.Furthermore, Figure \ref{fig:recall} records the precision and recall scores of all the speech tokens on 80-10-10 train-dev-test split. In Figure \ref{fig:datasplitph2}, we again vary the train-test ratio of data and present the performance accuracy for speech token prediction corresponding to the top 4 models as indicated in Table \ref{test_result2}.

\section{Conclusion and Contribution}
We report a novel hierarchical deep neural network architecture composed of parallel spatio-temporal CNN and a deep autoencoder for phonological and speech token prediction from imagined speech EEG data. Overall, we made the following contributions: (1) we proposed a novel method for embedding the high dimensional EEG data into a cross-covariance matrix that captures the joint variability of the electrodes. Rather than attempting to directly decode speech thoughts into speech tokens, (2) we exploited the cross-covariance matrix to successfully classify the phonological attributes of these thoughts into 6 categories; and (3) we used these predicted phonological categories to identify speech tokens. Ultimately, (4) our work suggests the existence of a brain imagery footprint for underlying ariculatory movements representing speech tokens.

\bibliographystyle{IEEEtran}

\bibliography{Saha_interspeech}

\end{document}